\documentclass[review,authoryear,11pt]{elsarticle}
\usepackage[margin=0.8in]{geometry}

\usepackage{graphics}
\usepackage{caption}
\usepackage{subcaption}

\usepackage{algorithm}
\usepackage{algorithmicx}
\usepackage{algpseudocode}
\usepackage{pifont}

\renewcommand{\vec}[1]{\mathbf{#1}}

\journal{Neural Networks}
\begin{document}
\begin{frontmatter}

\title{Multi-Output Artificial Neural Network for Storm Surge Prediction in North Carolina}

\author[MCSaddress]{Anton Bezuglov\corref{mycorrespondingauthor}}
\cortext[mycorrespondingauthor]{Corresponding author}
\ead{bezuglova@benedict.edu}

\author[RENCIaddress]{Brian Blanton}
\ead{bblanton@renci.org}

\author[MCSaddress]{Reinaldo Santiago}
\ead{santiagor@benedict.edu}

\address[MCSaddress]{Math and Computer Science Dept, Benedict College, 1600 Harden St., Columbia, SC, 29204}
\address[RENCIaddress]{Renaissance Computing Institute, The University of North Carolina at Chapel Hill, 100 Europa Dr., Chapel Hill, NC, 27517}

\begin{abstract}
During hurricane seasons, emergency managers and other decision makers need accurate and `on-time' information on potential storm surge impacts.  Fully dynamical computer models, such as the ADCIRC tide, storm surge, and wind-wave model take several hours to complete a forecast when configured at high spatial resolution.  Additionally, statically meaningful ensembles of high-resolution models (needed for uncertainty estimation) cannot easily be computed in near real-time. This paper discusses an artificial neural network model for storm surge prediction in North Carolina. The network model provides fast, real-time storm surge estimates at coastal locations in North Carolina. The paper studies the performance of the neural network model vs. other models on synthetic and real hurricane data. 
\end{abstract}


\end{frontmatter}
\section{Introduction}
\label{sctIntro}
During hurricane seasons, emergency managers and other decision makers need accurate and `on-time' information on potential storm surge impacts.  Fully dynamical computer models, such as the ADCIRC tide, storm surge, and wind-wave model \citep{Westerink_etal:08}, when configured at high spatial resolution, typically take several hours to complete a forecast. This complicates computing statically meaningful ensembles of high-resolution models (needed for uncertainty estimation) in near real-time. Contrary to this, an artificial neural network (ANN) can be an accurate and fast non-parametric model with computational complexity~\citep{skiena1998algorithm} of only $O(N^2)$ for a two layer feedforward neural network with N neurons in the hidden layer.
%

This paper focuses on design, implementation, and testing of an ANN model for storm surge prediction in North Carolina. The inputs to the model are hurricane parameters such as its location, central pressure, and radius to maximum winds. The outputs are the storm surge predictions for specified locations along the coast. We start with a short description of the storm surge simulation process and needed inputs, followed by review of machine learning techniques as applied to storm surge.  The network architecture is then described, followed by application to the North Carolina coast. The contributions of this work are as follows:
\begin{itemize}
\item Fast and accurate model for NC coastal areas -- the proposed ANN was tested on historic and simulated hurricane tracks;
\item Multiple outputs -- one ANN serves multiple locations;
\item Open source implementation -- the model is freely available and implemented using TensorFlow library \citep{abadi2015tensorflow};
\end{itemize}

Generally, storm surge dynamics is a relatively well-understood area in geosciences (see \cite{Flather:2001} for an excellent overview of the physics).  Recent research activities, motivated by major hurricane events such as Hurricanes Katrina (2005) and Sandy (2012), have focused on the ability to model and predict storm surge both for deterministic simulations~\citep{Lin_etal:2010a, Dietrich_etal:2011,Dresback_etal:2013} and for probabilistic/ensemble-based approaches that allow assessment of uncertainties~\citep{Davis_etal:10, di2011verification, bernier2015deterministic}.  Dynamic models for storm surge directly solve the shallow water wave equations given specified initial and boundary conditions~\citep{ResioWesterink:08}, and include NOAA’s SLOSH~\citep{Jelesnianski_etal:1992}, FVCOM~\citep{FVCOM_1}, and the Advanced Three-Dimensional Circulation Model (ADCIRC)~\citep{luettich1992adcirc,Westerink_etal:08}. For hurricane-driven storm surge simulations, the boundary conditions include specification of the atmospheric surface pressure and surface wind velocity that describe the time-evolution of the hurricane vortex. The source of the winds and pressures themselves can represent an historical event, a synthetic event, or a forecast. Regardless, the data are usually either from dynamic models for the atmosphere (e.g., the National Weather Service's operational forecast models) or parametric descriptions of the vortex in terms of a few parameters.   These parameters can be derived from analysis of the observed hurricane history as characterized in datasets such as the International Best Track Archive for Climate Stewardship (IBTrACS,~\cite{Knapp_etal:2010}) or actual hurricane forecasts from operational groups such as the National Hurricane Center.  The fundamental parameters that describe hurricanes are the pressure deficit between the center of rotation and the far-away ambient conditions, the distance from the center of rotation to the location of maximum wind speeds, and the forward speed of the vortex~\citep{holland1980analytic, holland2008revised, holland2010revised}.  Good overviews of hurricanes are given in ~\cite{ROG:ROG304} and ~\cite{marks2003hurricanes}.

\begin{figure}
\begin{center}
\centering 
\includegraphics[width=0.8\linewidth]{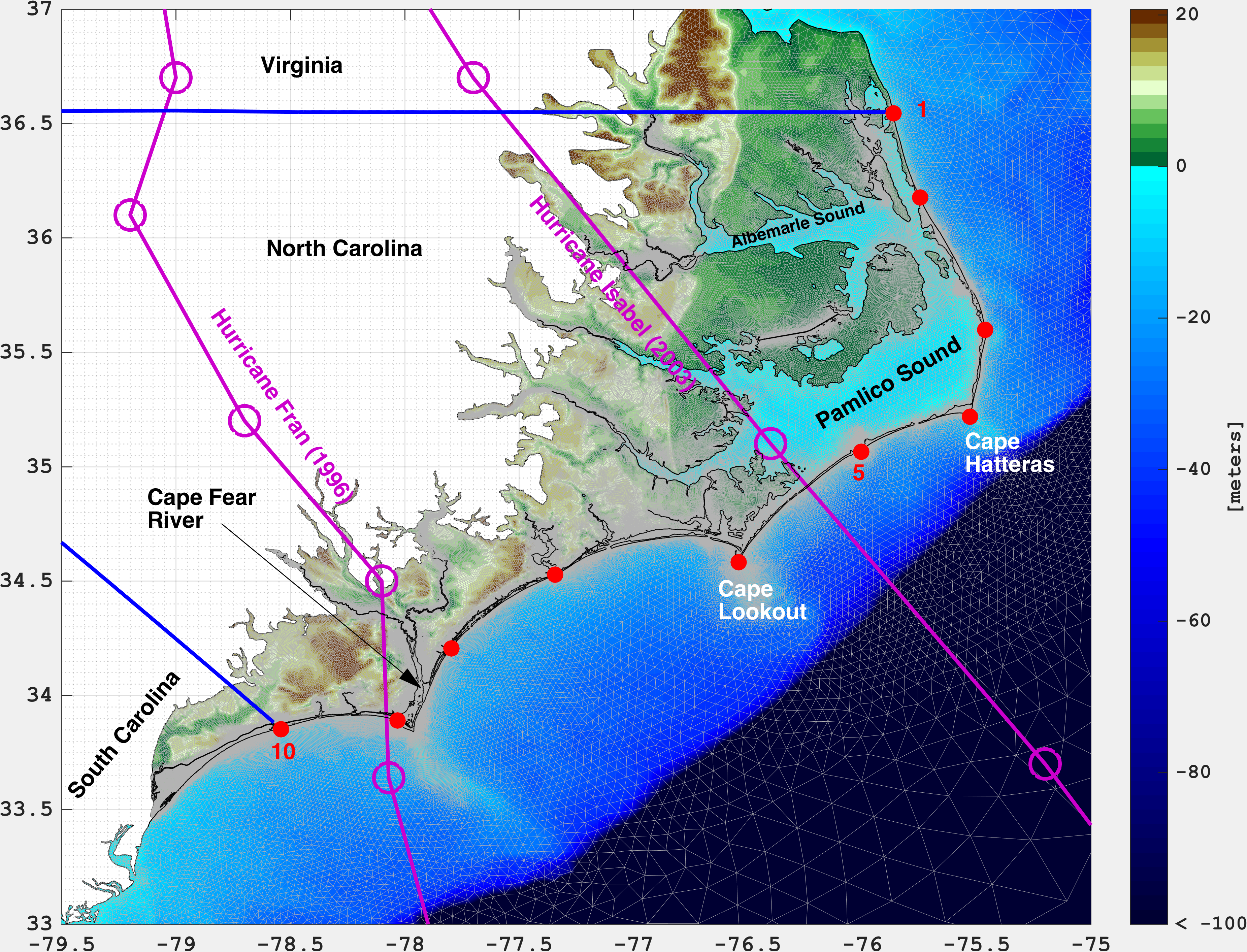}
\caption{ADCIRC's finite element grid for the North Carolina coast.  The linear, triangular elements are shown in gray, with the coastline drawn in black.  Colors represent the land topographic heights (green to brown) and ocean floor depths (blues). The high resolution is indicated by the density of triangles near the coast and along rivers.  The ten output locations are shown with red dots, numbered from north to south. The paths of Hurricanes Fran (1996) and Isabel (2003) are shown with purple lines.}
\label{Fig:adc_grid}
\end{center}
\end{figure}

Dynamic models like ADCIRC can be configured at a very high spatial resolution (Figure~\ref{Fig:adc_grid}), but at a concomitant high cost in terms of computer resources needed to conduct a simulation in near-real-time ~\citep{Blanton_etal:2012}.  To overcome this challenge, surrogate models can be developed that leverage pre-computed datasets of synthetic hurricane parameters and paths and storm surge responses.  Regression models approximate the (unknown) functional relationship between inputs (hurricane parameters) and responses (storm surge), and include  
response surface methods~\citep{WICS:WICS73} applied to rapid storm surge simulations ~\citep{taflanidis2013rapid}, as well as kriging~\citep{jia2013kriging}.  

%
%
%
%
Non-parametric models are largely based on nonlinear regressions, support vector machines (SVM), and ANNs. 
The support vector method for function regression was introduced in~\cite{vapnik1997support} and \cite{smola1997support}. The core of the method is the proposition that a function is approximated by a linear combination of a small number of support vectors. Support vector regression is used for building non-parametric models.~\citep{yu2006support, zhang2006support, sreekumar2015one, wolff2016comparing, ma2003accurate, khemchandani2009regularized}.  \cite{rajasekaran2008support} applied support vector methodology for storm surge prediction and verified it using observations of water level at the Longdong station at Taiwan. 

ANNs belong to a family of non-parametric models and have a wide area of application~\citep{zhang1998forecasting, govindaraju2013artificial, aminzadeh2013geophysical}, including  storm surge predictions. \cite{bajo2010storm} developed an ANN to reduce the error in storm surge predictions made by a parametric hydrodynamic model for Venice, Italy.  \cite{lee2006neural,lee2008back} focused on two layer ANNs with the number of neurons ranging from 3 to 12 for typhoon storm surge. The models were trained using data from three typhoons  impacting Taiwan in 2003, including Typhoon Vamco, at three locations and produced outputs with root mean squared error (RMSE) from $0.05$ to $0.11$ m and correlation coefficients ($R$) from $0.97$ to $0.99$.
\cite{de2009neural} developed a two layer ANN for south east coastal region of Brazil. The model was trained on meteorological parameters (zonal wind stress, meridian wind stress, pressure, wind speed) and filtered sea level series of previous hours. The authors observed that the network with 11 neurons in the hidden layer produced the best results ($R$ from 0.851 to 0.998). 

One problem with applying non-parametric models, and ANNs in particular, is insufficient amounts of the data for model fitting. To overcome this, \cite{kim2015time} generated a set of 446 synthetic hurricanes with corresponding storm surge levels computed with the ADCIRC model. The dataset was used to train a two-layer model with number of neurons ranging from 16 to 25. The model was tested on historical hurricanes Katrina and Gustav and demonstrated $R$ from 0.917 to 0.996. A similar approach was proposed by~\cite{hashemi2016efficient}, where they used 1050 synthetic tropical storms to build a model for Rhode Island coastal waters. \cite{das2011efficient} used 47 synthetic storms to construct their storm surge forecasting tool for coastal Mississippi. The authors posited that a more comprehensive set of storms and storm surge responses would improve the results of the method.

In this paper, we describe extensions of the above applications.  First, we implement a multi-output prediction model, where a single ANN model produces storm surge predictions at multiple (currently ten) locations along the coast.  Second, we use a three layer model (two hidden layers) where the first hidden layer pre-processes hurricane data and the second hidden layer calculates individual predictions.  The rest of the paper is organized as follows: Section~\ref{sctArchitecture} reviews basics of ANNs, discusses dealing with multiple locations and implementation of the model, Section~\ref{sctData} overviews the data used in the study, Sections~\ref{sctExperiments} and~\ref{sctConclusions} discuss the findings and conclusions.
\section{Feedforward Artificial Neural Network (FF ANN) for Storm Surge Prediction}
\label{sctArchitecture}
\subsection{Model Architecture}
We briefly describe the application of artificial neural networks for storm surge prediction. More general and thorough discussions of the networks and applicable algorithms can be found in~\cite{bishop2006pattern} and \cite{de2009neural}, among many others.  Suppose hurricane data is described by a vector $\vec{x}$ of parameters such as longitude and latitude of its center, its forward or translation speed, radius from the center of rotation to maximum winds, etc. Suppose also that vector~$\vec{y}$ represents storm surge predictions at several locations. Assuming that $\vec{y}$ does not depend on factors other than $\vec{x}$, it can be evaluated as $\vec{y} = f(\vec{x})$, where $f(\cdot)$ is some modeling function. Depending on the implementation, $f(\cdot)$ can take various forms and in case of a two layer feed-forward ANN it is:
\begin{eqnarray*}
f(\vec{x}) = W_h*\vec{h}+\vec{b_h}\\
\vec{h} = \sigma(W_i*\vec{x}+\vec{b_i})
\label{eqn:simple_model}
\end{eqnarray*}
where $\vec{h}$ is the output of the hidden layer, and $W_h$, $W_i$, $\vec{b_h}$, $\vec{b_i}$ are weight matrices and biases for the the hidden and input layers respectively. The weight matrices and biases constitute a set of independent network parameters. $\sigma(\cdot)$ is a nonlinear function applied component-wise to the vector, which is essential as it allows the network to devise a presentation of a nonlinear function. $\sigma(\cdot)$ is typically a hyperbolic tangent or a sigmoid function:
%
\begin{eqnarray*}
\sigma_{tanh}(x) = \frac{2e^{-x}-1}{2e^{-x}+1}
\label{eqn:nonlinear_tanh}\\
\sigma_{sigmoid}(x) = \frac{1}{1+e^{-x}}
\label{eqn:nonlinear_sigm}
\end{eqnarray*}

\begin{figure}
\begin{center}
\centering 
\includegraphics{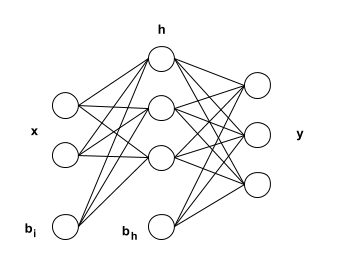}
\caption{Architecture of a two layer ANN}
\label{Fig:two_layer_ANN}
\end{center}
\end{figure}

Figure~\ref{Fig:two_layer_ANN} illustrates the architecture of a two-layer ANN. The information in the network travels from left to right (i.e., from the inputs $\vec{x}$ through a series of operations to the outputs $\vec{y}$). The vertices represent input, output, hidden variables, and placeholders for biases, and the edges represent independent parameters (i.e., the weight matrices and the biases themselves).

Suppose for each $\vec{x}$, the observations of `true' storm surges $\vec{\hat{y}}$ are known. Then, the model can be \emph{trained} to minimize the error between $\vec{\hat{y}}$ and $\vec{y}$ by changing its independent parameters. \cite{rumelhart1988learning} developed a highly efficient iterative back-propagation algorithm which is now commonly used for training. At each step, the algorithm calculates gradients for each model parameter to minimize the error from the outputs back to the inputs. The parameters are changed in accordance to their gradients and the procedure repeats until convergence.

The complexity of ANN algorithms is determined by the number of elementary operations, in this case multiplications. The number of multiplications follows from the size of the weight matrices and as such depends on the number of neurons in the hidden layer. In case of a fully trained ANN, the network's inference complexity is expected to be $O(N^2)$, where $N$-number of neurons. The network training complexity will then be~$O(N^2*k*E)$, where $k$-training dataset size and $E$-number of epochs (iterations).

\subsection{Dealing with Multiple Outputs}
\label{subsctMultipleOutputs}
In contrast to previous studies, here the output vector $\vec{y}$ contains 10 components, one for each location along the coast. A straight-forward approach to modeling would be to have 10 separate two-layer networks one per each output and train them individually. 
%
%
This has been thoroughly tested and the literature suggests anywhere between eight and thirty neurons in the hidden layer, depending on the complexity of geography and other factors. However, due to spatial correlation between the outputs, the networks will be somewhat similar and redundant. Besides, intuitively, each hurricane track is a \emph{single} event that impacts multiple locations and as such the information has to be shared across locations. 

We have used a single network with multiple outputs to mitigate the impacts of this  redundancy. Additionally, to improve information sharing in the network, the second hidden layer was included resulting in a three layer multiple output ANN. It is difficult, if not impossible, to make propositions regarding internals of ANNs, but intuition and empirical evidence suggest that the first hidden layer serves as an `input preparation' layer, converting $\vec{x}$ to a better form, whereas the second hidden layer `localizes' the outputs. Another proposition is to have more neurons in the second hidden layer compared to the first, since the output has more components than the input.  
\begin{table}[!htb]
\caption{Preliminary analysis of network types}
\label{tab:architectures}     
\centering
\begin{tabular}{l | c | c }
\hline\noalign{\smallskip}
ANN type & MSE & $R$\\
\noalign{\smallskip}\hline\noalign{\smallskip}
10~independent two-layer ANNs (32) & 0.010 & 0.90-0.95 \\
Two layer (200) & 0.014-0.025 & 0.92-0.95\\
Three layer (32,32) & 0.007 & 0.92-0.96 \\
Three layer (64,32) & 0.008 & 0.90-0.97\\
Three layer (64,64) & 0.008 & 0.92-0.98\\
Three layer (32,64) & 0.006-0.008 & 0.93-0.98 \\
\noalign{\smallskip}\hline\noalign{\smallskip}
\end{tabular}
\end{table}

Table~\ref{tab:architectures} summarizes the empirical data for each network type. The networks are compared by using mean squared errors (MSE) and average correlation coefficients ($R$), which are more formally given in Section~\ref{sctExperiments}. The first row of the table summarizes the average performance of 10 independent two-layer ANNs with 32 neurons in the hidden layer. The second row gives the performance of a two-layer multiple output network with 200 neurons. The following rows give the performances of three layer multi-output networks with various hidden layer sizes. According to the table, the three layer multiple output networks are appropriately accurate and simple, which substantiates the propositions above. 

\subsection{Implementation}
Model equations for the multiple output three layer network are as follows:
\begin{eqnarray*}
f(\vec{x}) = W_{o}*\vec{h_2}+\vec{b_{o}}\\
\vec{h_2} = \sigma_{tanh}(W_{h2}*\vec{h_1}+\vec{b_{h2}})\\
\vec{h_1} = \sigma_{tanh}(W_{h1}*\vec{x}+\vec{b_{h1}})
\label{eqn:simple_model}
\end{eqnarray*}
where $\vec{h_1}$ and $\vec{h_2}$ are the outputs of the first and second hidden layers; $W_{h1}$, $W_{h2}$, $W_{o}$, $\vec{b_{h1}}$, $\vec{b_{h2}}$, $\vec{b_{o}}$ are weight matrices and biases for the first, second and output layers respectively. The model is trained using the back-propagation algorithm to minimize MSE of the training dataset of size $N$ for $K=10$ outputs:
\begin{eqnarray*}
MSE = \frac{1}{KN}\sum_{i=1}^{K}\sum_{j=1}^{N}\left(f(x_i^j)-\hat{y}_i^j\right)^2\\
\label{eqn:simple_model}
\end{eqnarray*}

The neural network was implemented and trained using Tensorflow library~\citep{abadi2015tensorflow}. The minimization of MSE was performed with the ADAM stochastic optimization algorithm~\citep{kingma2014adam} with batch sizes from 19 to 57 tracks. The code can run on systems with multiple GPUs, in which case it calculates update gradients in parallel. Algorithm~\ref{alg:multi_gpu_training} summarizes the implementation and the complete code is available to download from~\citep{git_url}.

\begin{algorithm}[!htb]
    \caption{Training algorithm for a multiple output ANN on parallel GPUs using TensorFlow}
    \label{alg:multi_gpu_training}
    \begin{algorithmic}[1] 
        \Procedure{Train}{$\vec{x},\vec{\hat{y}}$} \Comment{$\vec{x}$ -- inputs,$\vec{\hat{y}}$ -- observations}
			\State $\vec{x_n}$ = Normalize($\vec{x}$) \Comment{subtract means and divide by std component-wise}
            \For{g in GPUs}
            	\State $g \gets InitializeNetwork()$ \Comment{Assign a copy of the network on each GPU}
		    \EndFor
            \For{e in Epochs}
   	        	\State $b_{1..GPUs} \gets GetDataBatch(\vec{x_n}, \vec{\hat{y}}, size)$ \Comment{Fetch batches of training data for each GPU}
                \State $g_{1..GPUs} \gets CalculateGradients(b_{1..GPUs})$ \Comment{Calculate gradients on each GPU}
            	\State $g \gets Mean(g_{1..GPUs})$ \Comment{Find average gradient}
                \State $\vec{v} \gets AdamOptimizer(g, rate)$ \Comment{Obtain the new values of network parameters}
                \State UpdateNetworks($\vec{v}$) \Comment{Propagate changes to all networks}
		    \EndFor
        \EndProcedure
    \end{algorithmic}
\end{algorithm}

\section{Dataset}
\label{sctData}
\begin{figure}[htb!]
\centering
\begin{subfigure}{.45\textwidth}
\centering
  \includegraphics[width=1.0\linewidth]{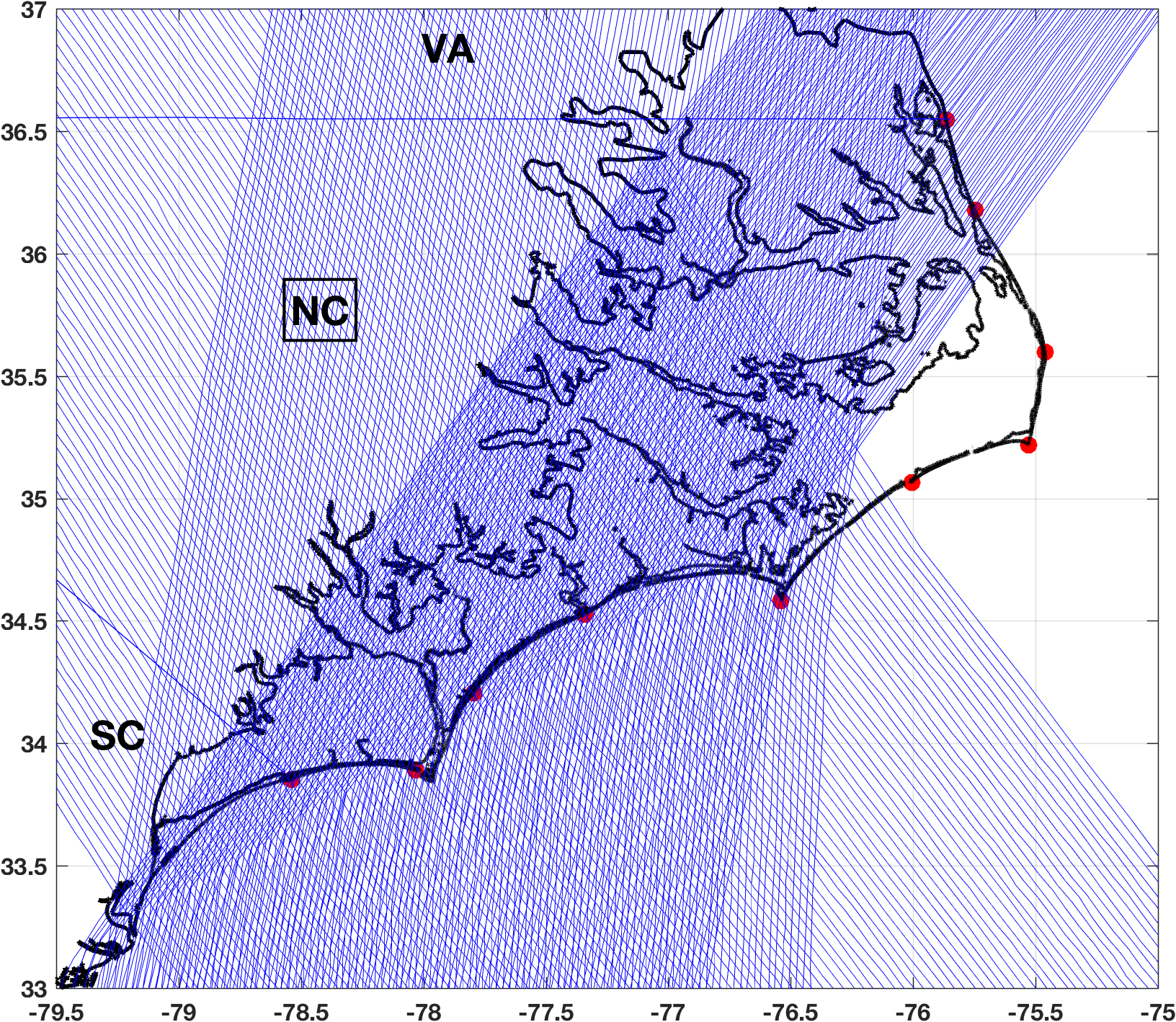}
\caption{}
\label{Fig:tracks} 
\end{subfigure}
\begin{subfigure}{.45\textwidth}
 \centering
\includegraphics[width=1.0\linewidth]{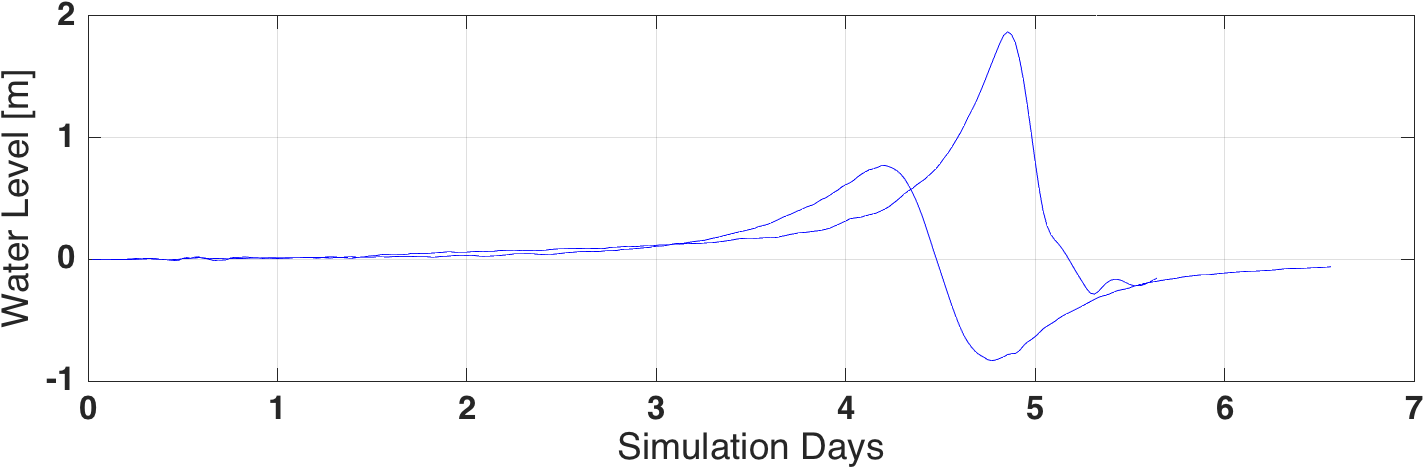}
  \caption{}
  \label{Fig:examp}
\end{subfigure}
\caption{a) Landfalling hurricane tracks derived from the historical record for the North Carolina coastal flood insurance study. There are 324 tracks. b) ADCIRC-simulated water levels [meters above mean sea level] for a location near Wrightsville Beach, North Carolina for two of the landfalling tracks.}
\label{Fig:examp}
\end{figure}

In this work, the ANNs were trained and validated on a dataset of synthetic hurricanes and storm surge responses for coastal North Carolina. The dataset was computed for a recent and detailed coastal flood insurance study for FEMA~\citep{Blanton2012ncfmpstats}. The hurricane tracks in the dataset are shown in Figure~\ref{Fig:tracks}, along with representative storm surge responses from ADCIRC (Figure ~\ref{Fig:examp}). There are a total of 324 storm events.  This set of hurricanes represents the statistics of landfalling hurricanes in North Carolina, including the landfall locations. Note that hurricane landfalls have not occurred north of Cape Hatteras (see~\ref{Fig:adc_grid}), although storms making landfall along the southeast-facing shore have transited across the coastal lands to exit back over water along the northern coast. We also do not include astronomical tides in the data set, since the hurricane events would occur at a random phase of the tide.

The hurricane parameters were used to train and validate the models. Each track is essentially a table with 193 rows and 16 columns. The first six columns represent $\vec{x}$: time to landfall in days, hurricane center coordinates, radius to maximum winds, maximum wind speed, and storm forward/translation speed. The duration of each track is four days, with 3 days prior to landfall and one day post-landfall. The last ten columns are the outputs $\vec{\hat{y}}$, i.e. storm surge in meters above mean sea level at the specified output locations (Figure~\ref{Fig:tracks}).


While discussing the dataset, it is important to touch on the subject of \emph{over-fitting}, when a model, in attempt to minimize the error on a training dataset, fails to generalize on previously unseen data. Over-fitting is typically a result of an overly complex model trained on an insufficiently large dataset. In this work, several methods were applied to mitigate over-fitting. First, a somewhat heuristic observation suggests that the ratio of data samples to free model parameters has to be approximately 4:1 or higher. If the number of free parameters in ANN is roughly the squared number of neurons, the recommended network size should be 125 neurons ($\sqrt{62532/4}$) or fewer. 

Second, the dataset is split into three subsets: training (70\%, 228 tracks), validation (15\%, 48 tracks), and testing (15\%, 48 tracks). The training subset is used to update the network by calculating gradients and adjusting network weights and biases. The validation subset is used to evaluate model performance, adjust hyper-parameters such as its learning rate, and to detect over-fitting. The model performance is evaluated on the testing dataset.

\section{Experiments and Discussion}
\label{sctExperiments}
Two metrics were applied to measure network performance: mean squared error (MSE) and correlation coefficient ($R$). Suppose that $y_i^j$ and $\hat{y}_i^j$ are model predictions and observations (ADCIRC) of storm surges at location $i$ for sample $j$, then 
\begin{eqnarray*}
MSE_i = \frac{1}{N}\sum_{j=1}^{N}\left(y_i^j-\hat{y}_i^j\right)^2\\
R_i = \frac{\sum_{j=1}^{N}(y_i^j-\bar{y}_i^j)(\hat{y}_i^j-\bar{\hat{y}}_i^j)}{\sqrt[]{\sum_{j=1}^{N}(y_i^j-\bar{y}_i^j)^2\sum_{j=1}^{N}(\hat{y}_i^j-\bar{\hat{y}}_i^j)^2}}
\label{eqn:simple_model}
\end{eqnarray*}

\noindent All tests below were performed on either synthetic testing datasets or historic hurricanes, which were not directly\footnote{The historic hurricanes were used to derive parameters and generate synthetic hurricanes} used for model training. 
\subsection{Synthetic Hurricanes}
Table~\ref{tab:synthetic_performance_overall} compares performance of several ANNs at each output location on the test set of 48~synthetic storms. For each network, the table shows the size of the hidden layer(s) and the number of epochs in the training. MSE column is the average mean squared error across all locations, and $R_i$ is the correlation coefficient at location $i$. The upper part of the table summarizes two-layer networks with 60, 120, and 200 neurons, and the lower part -- the three layer networks. For each network, multiple training sessions were needed to identify the best hyper-parameters such as the learning rate, number of epochs, learning rate decay, batch size and others. The table summarizes the best performances of each network type.

\begin{table}[!htb]
\caption{Performance on synthetic hurricane tracks.  Correlation coefficients $R$ are given for output locations, numbered 1-10 from north to south as shown in Figure \ref{Fig:adc_grid}.}
\label{tab:synthetic_performance_overall}     
\centering
\begin{tabular}{c | c | c | c | c | c | c | c | c | c | c | c | c }
\hline\noalign{\smallskip}
ANN ($N_1$, $N_2$) & Epochs & MSE & $R_1$ & $R_2$ & $R_3$ & $R_4$ & $R_5$ & $R_6$ & $R_7$ & $R_8$ & $R_9$ & $R_{10}$\\
\noalign{\smallskip}\hline\noalign{\smallskip}
(60,-) &10,000& 0.014 & 0.90 & 0.91 & 0.92 & 0.93 & 0.92 & 0.89 & 0.94 & 0.93 & 0.85 & 0.89 \\
(120,-) &10,000& 0.012 & 0.91 & 0.92 & 0.93 & 0.95 & 0.93 & 0.88 & 0.92 & 0.91 & 0.87 & 0.89 \\
(200,-) &25,000& 0.014 & 0.93 & 0.94 & 0.94 & 0.95 & 0.95 & 0.92 & 0.95 & 0.94 & 0.92 & 0.93 \\
\noalign{\smallskip}\hline\noalign{\smallskip}
(16,32) &15,000& 0.007 & 0.95 & 0.96 & 0.97 & 0.97 & 0.97 & 0.95 & 0.97 & 0.96 & 0.93 & 0.93 \\
(32,32) &15,000& 0.007 & 0.95 & 0.96 & 0.96 & 0.96 & 0.96 & 0.93 & 0.96 & 0.96 & 0.94 & 0.94 \\
(16,64) &15,000& 0.006 & 0.97 & 0.97 & 0.97 & 0.98 & 0.98 & 0.93 & 0.97 & 0.96 & 0.95 & 0.95 \\
(32,64)* &15,000& 0.006 & 0.96 & 0.96 & 0.97 & 0.98 & 0.97 & 0.95 & 0.97 & 0.98 & 0.95 & 0.95\\
(64,128) &15,000& 0.006 & 0.96 & 0.96 & 0.97 & 0.98 & 0.98 & 0.94 & 0.96 & 0.96 & 0.93 & 0.92 \\
\noalign{\smallskip}\hline\noalign{\smallskip}
\end{tabular}
\end{table}

The table suggests a few observations. First, locations 6, 9, and 10 are more challenging for all the networks, whereas locations 3, 4, 5, and 7 appear to be the `easiest', which is explained by the shape of the coast. Second, the MSEs of three layer networks are about half that of the two layer networks, which further justifies the propositions given in Section~\ref{subsctMultipleOutputs}. Third, $R$ and MSE consistently improve with the growth of network complexity until it reaches (64,128). No further improvements are likely because of over-fitting. The (32,64) network demonstrates a good balance between accuracy and complexity and so it was selected for further analysis.

\begin{figure}[!htb]
\centering
\begin{subfigure}{.45\textwidth}
\centering
	\includegraphics[width=1.0\linewidth]{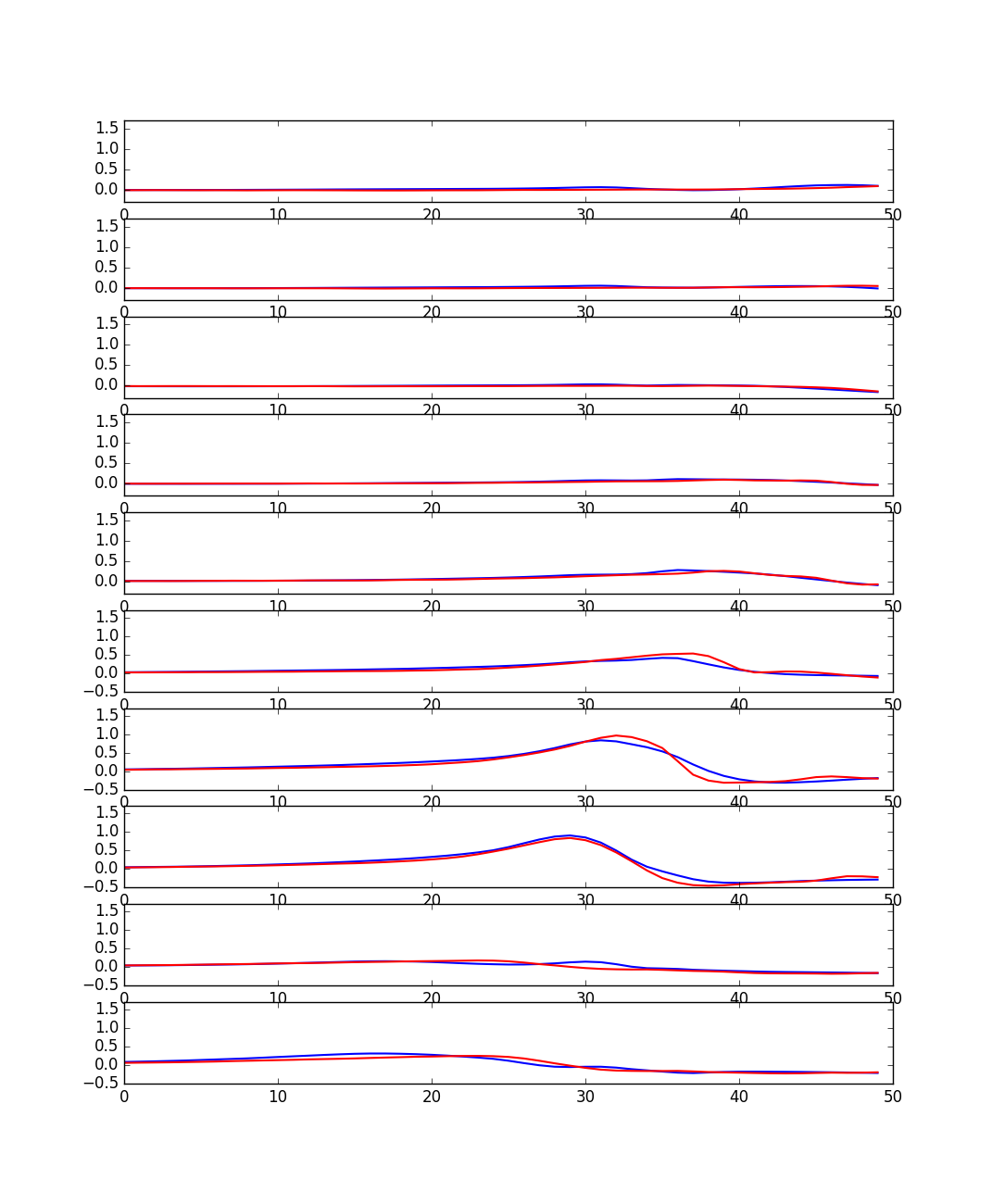}
	\caption{Track 1}
\end{subfigure}
\begin{subfigure}{.45\textwidth}
\centering
	\includegraphics[width=1.0\linewidth]{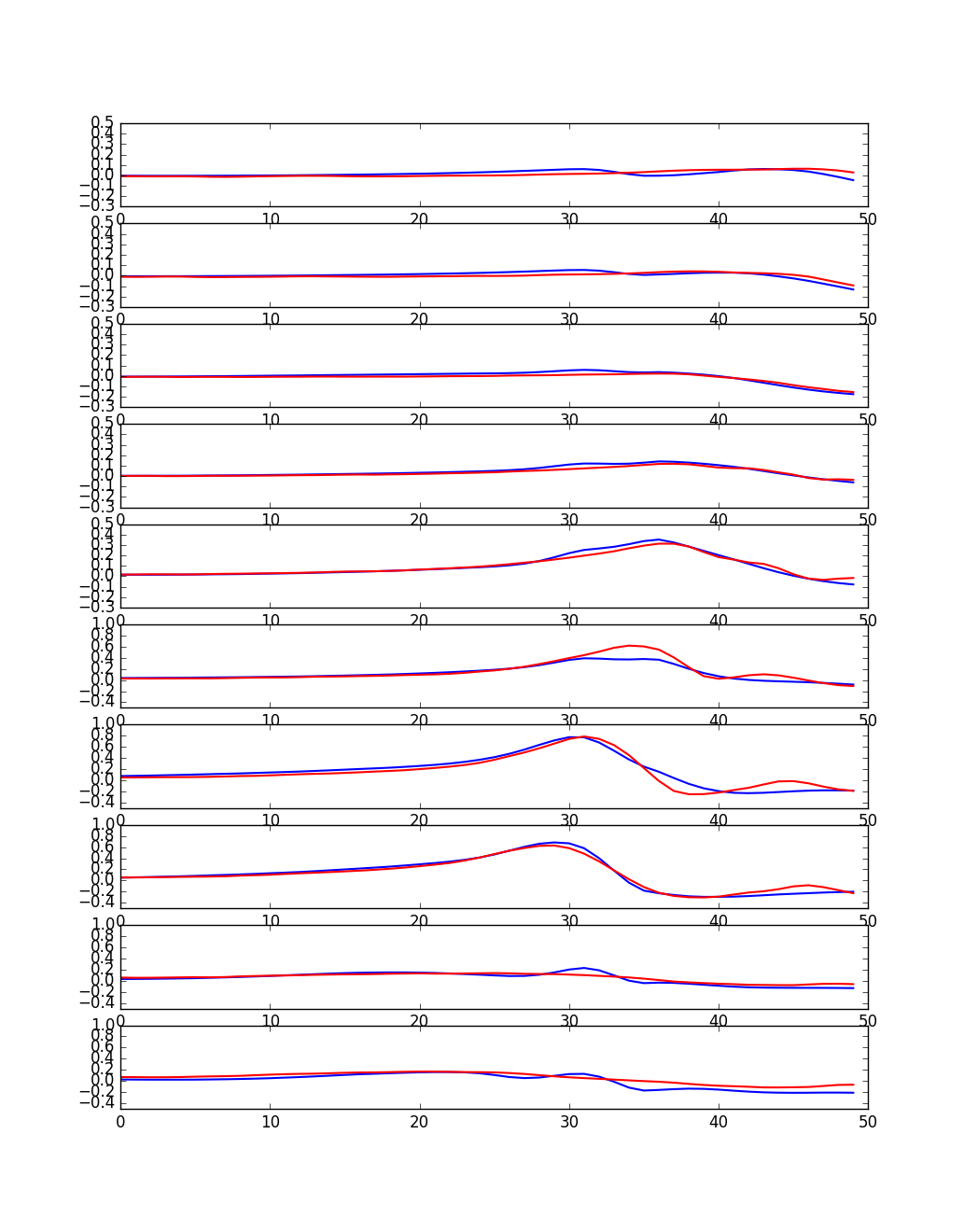}
	\caption{Track 2}
\end{subfigure}
\begin{subfigure}{.45\textwidth}
\centering
	\includegraphics[width=1.0\linewidth]{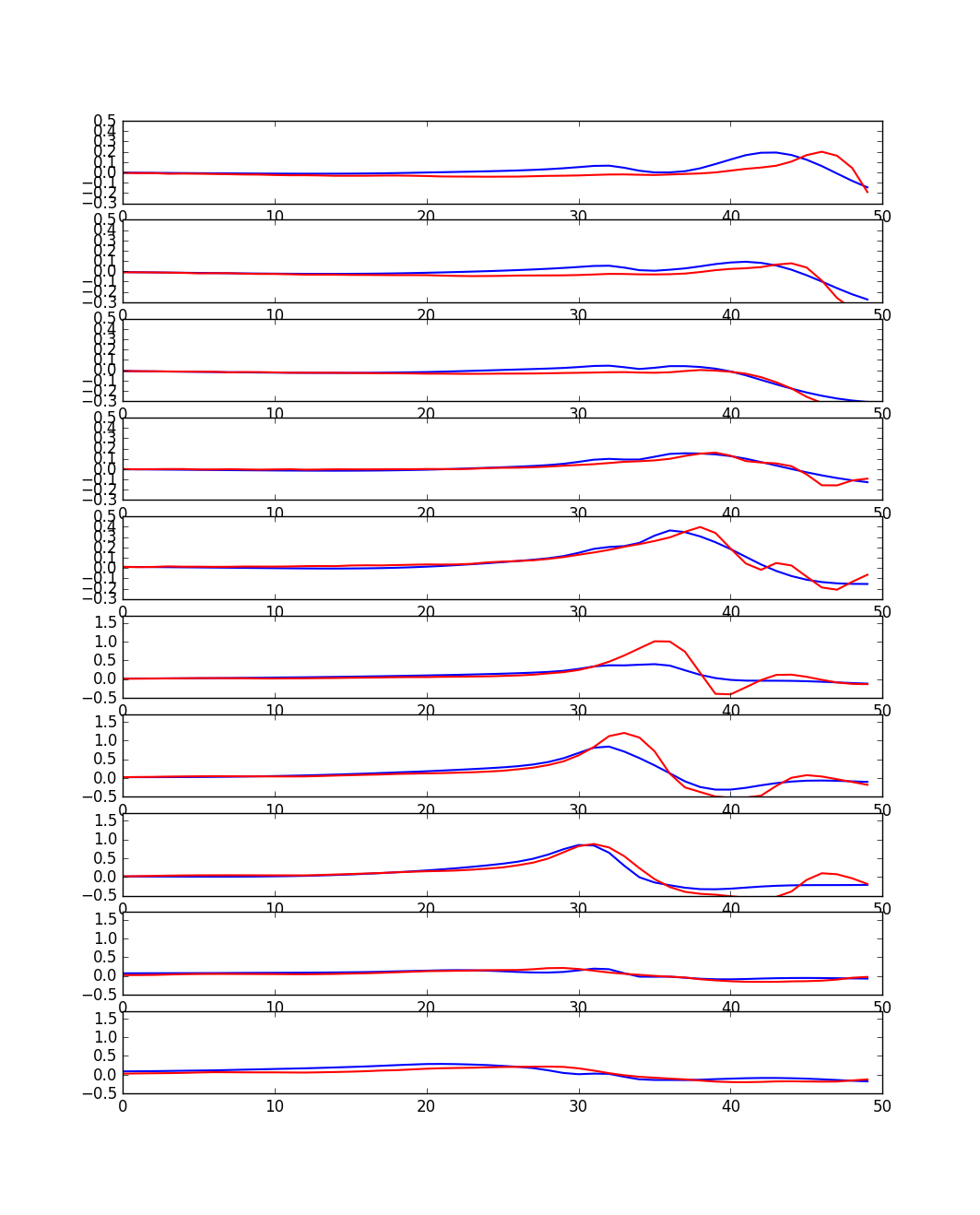}
	\caption{Track 3}
\end{subfigure}
\begin{subfigure}{.45\textwidth}
\centering
	\includegraphics[width=1.0\linewidth]{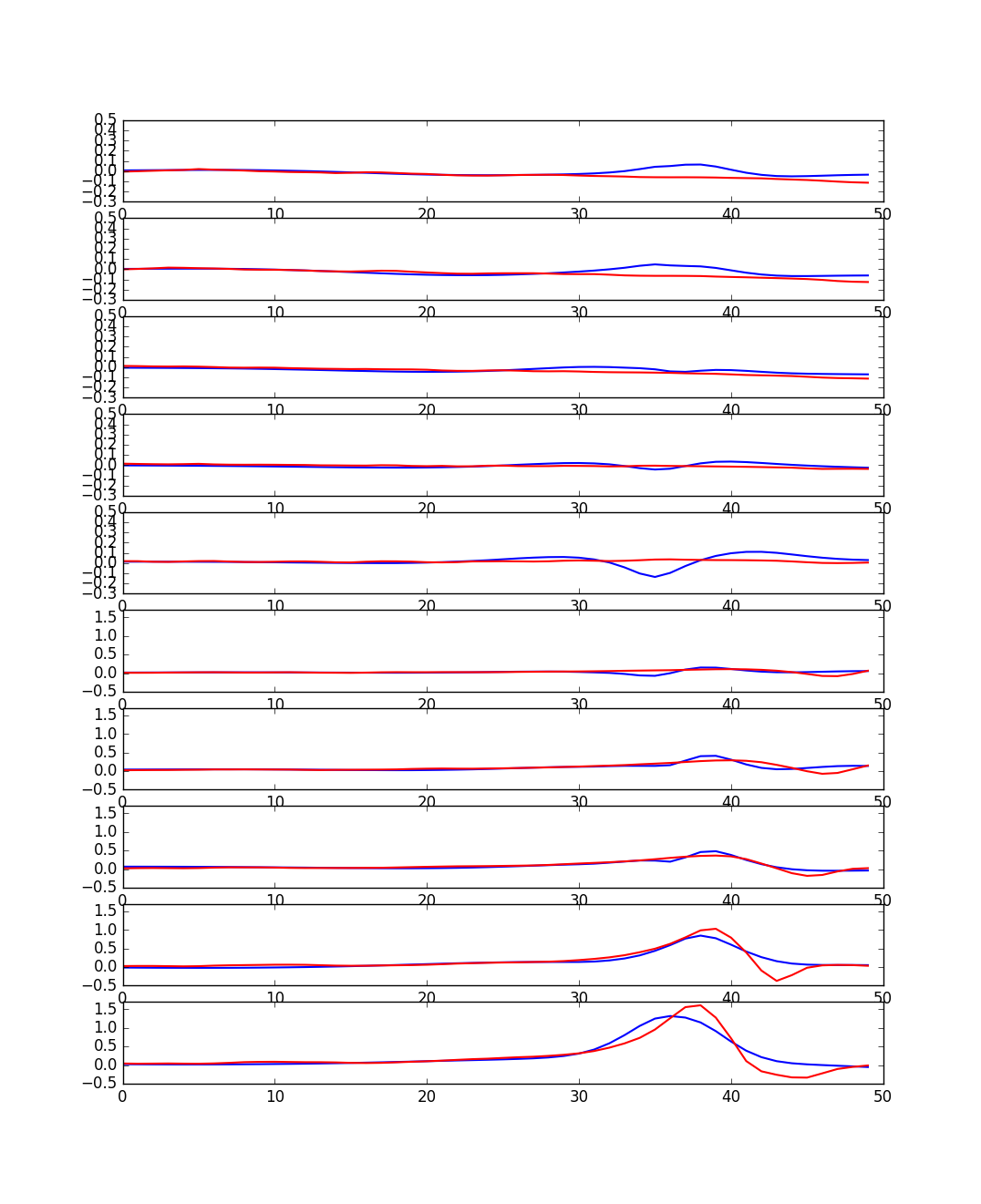}
	\caption{Track 4}
\end{subfigure}
\caption{Performance of ANN (32, 64) for four synthetic hurricane tracks. ADCIRC model results ("observations") are shown with red lines; ANN model outputs are shown with blue lines.}
\label{Fig:32_64_performance} 
\end{figure}

Figure~\ref{Fig:32_64_performance} demonstrates outputs of the (32,64) network on a few synthetic hurricane tracks. Each sub-figure shows the ten locations arranged vertically, from north (top) to south (bottom). As seen in the figure, tracks 1 and 2 were modeled accurately, whereas in track 3, the observations at locations 6 and 7 were underestimated. An interesting property of the network can be seen at location 5 for tracks 1 and 2. Even though the outputs at other locations were somewhat similar, the shared information from them did not inadvertently  cause detrimental effects on the predictions at location 5.

\subsection{Analysis of network errors}
As demonstrated by Table~\ref{tab:synthetic_performance_overall}, the mean prediction errors of the networks are low and are on average between~7 and~8 centimeters. Unfortunately, both MSE and $R$ can be substantially influenced by accurate predictions \emph{before} and \emph{after} the landfalls and thus neither may represent the acceptable accuracy during times of peak water level. This section considers prediction errors by ANN (32,64), which demonstrated the best results previously. 

First, the ANN ran on the complete set of synthetic hurricane tracks and the errors were calculated as a difference between network outputs and observations for each data sample at each location. Assuming that at each location, the errors are independent, their probability density function (PDF) was fitted using kernel density estimation~\citep{scott2015multivariate}. This approach provides an `optimistic' view on model errors as it still includes error-free regions long before the landfalls.

In order to get a realistic estimation of network accuracy, the further analysis only included regions immediately prior and after the landfall that encompass the largest absolute errors. Identically to the previous case, for each location, error PDF was fitted and error probabilities evaluated. Table~\ref{tab:model_errors} summarizes the results for the first and the second cases.
\begin{table}[!htb]
\caption{Accuracy of (32,64) network on synthetic data}
\label{tab:model_errors}     
\centering
\begin{tabular}{ c | c | c | c | c | c | c }
\hline\noalign{\smallskip}
Location & MSE & $R$ & $P(|e|\leq 0.1m)$ & $e^*, P(|e|\leq e^*)=0.95$ & $P(|e|\leq 0.1m)$ & $P(|e|\leq 0.5m)$ \\
\noalign{\smallskip}\hline\noalign{\smallskip}
1 & 0.0017 & 0.963 & 0.966 & $0.15$ & 0.864 & 0.998 \\
2 & 0.0012 & 0.975 & 0.976 & $0.13$ & 0.909 & 0.998 \\
3 & 0.0008 & 0.988 & 0.992 & $0.10$ & 0.950 & 0.999 \\
4 & 0.0004 & 0.992 & 0.993 & $0.10$ & 0.953 & 0.999 \\
5 & 0.0014 & 0.976 & 0.978 & $0.17$ & 0.861 & 0.994 \\
6 & 0.0038 & 0.932 & 0.935 & $0.32$ & 0.692 & 0.985 \\
7 & 0.0079 & 0.892 & 0.900 & $0.46$ & 0.531 & 0.960 \\
8 & 0.0112 & 0.853 & 0.864 & $0.51$ & 0.477 & 0.949 \\
9 & 0.0095 & 0.901 & 0.910 & $0.44$ & 0.566 & 0.960 \\
10 & 0.0175 & 0.833 & 0.850 & $0.49$ & 0.475 & 0.953 \\
\noalign{\smallskip}\hline\noalign{\smallskip}
\end{tabular}
\end{table}

In the table, the second column gives per location probabilities of the error to be within 10 centimeters for all the data, including pre-landfall regions. The last three columns give intervals and probabilities for the time window surrounding landfall. 

\subsection{Observed hurricane performance}
Since the longterm goal is to develop prediction methods that provide usable information for decision makers, we used the ANN (32,64) model to predict the storm surge (without astronomical tides) for two observed hurricanes that impacted North Carolina, Hurricane Fran (1996) and Hurricane Isabel (2003).  The storm tracks are shown in Figure~\ref{Fig:adc_grid}.  The input hurricane parameters for ANN were extracted from the IBTrACS dataset and interpolated to match the time intervals in the training data (30 minutes).  For each hurricane, we computed a "truth" solution using the ADCIRC model and using the same model configuration as that used to compute the training dataset.  The parameters for neither storm are explicitly in the data set; however, they are in the historical data from which the data set parameters were derived.  

\begin{figure}
\centering
\includegraphics[scale=0.75]{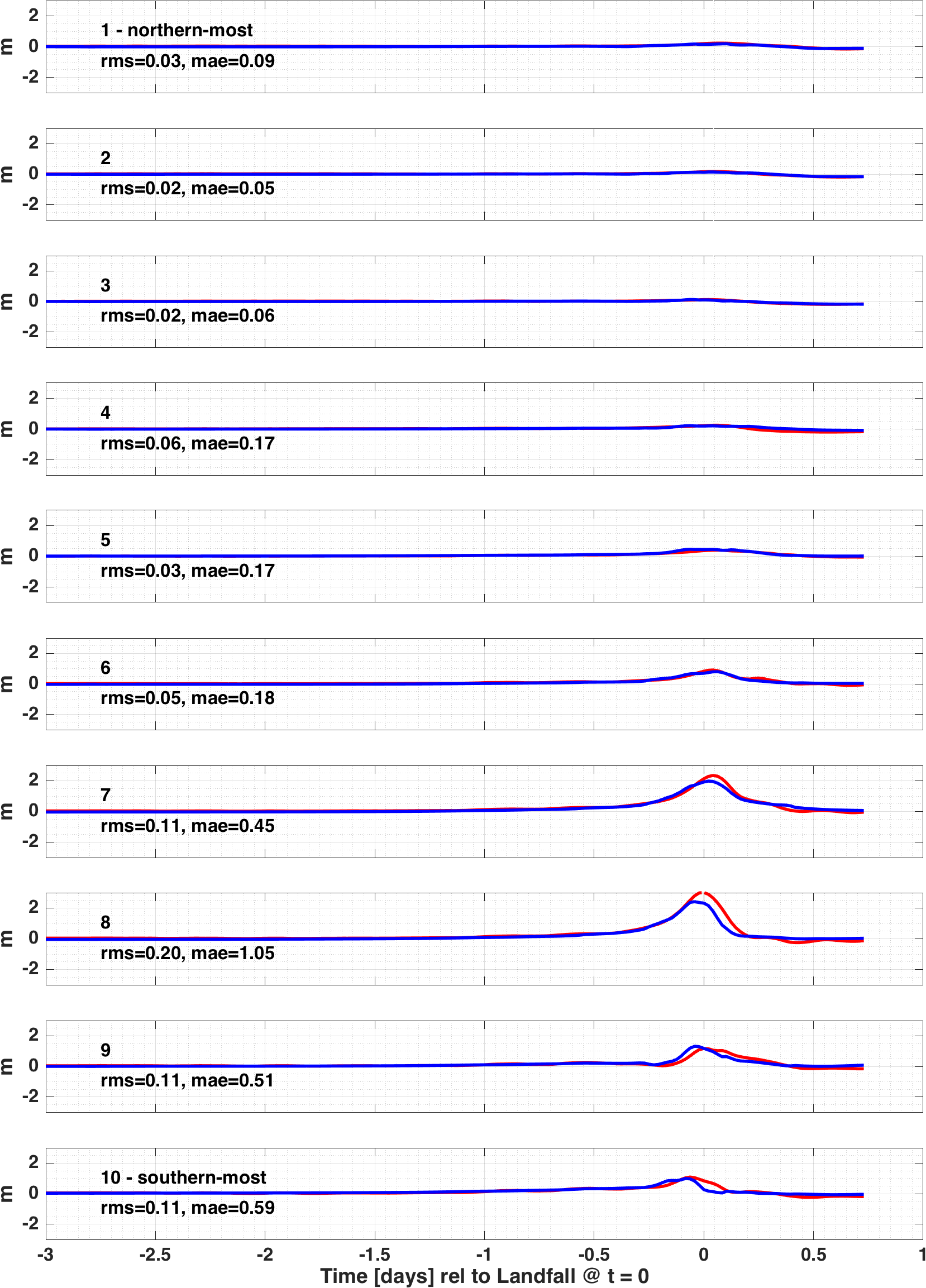}
\caption{Performance of ANN (32, 64) on Hurricane Fran (1996). ADCIRC model results ("observations") are shown with red lines; ANN model outputs are shown with blue lines.}
\label{Fig:Fran_performance} 
\end{figure}

Hurricane Fran made landfall near Southport, NC, (output location 9 on Figure~\ref{Fig:adc_grid}) and caused significant flooding and coastal high water that approached 4 m (~13 ft) in the Wrightsville Beach area (output location 8).  Figure~\ref{Fig:Fran_performance} shows the ANN response at the 10 output locations, along with the "truth" response.  In the figure, the red lines represent ADCIRC and blue lines are the ANN model outputs. The figure also provides root mean squared errors (rms) and mean absolute errors (mae) in meters for each coastal location.  Fran had very little coastal impact north of Cape Lookout, noted by the very flat response at the northern locations 1-5.  The peak storm surge occurs at location 8, where the ANN and ADCIRC values are 2.25 and 2.6 m, respectively. 

\begin{figure}
\centering
\includegraphics[scale=0.75]{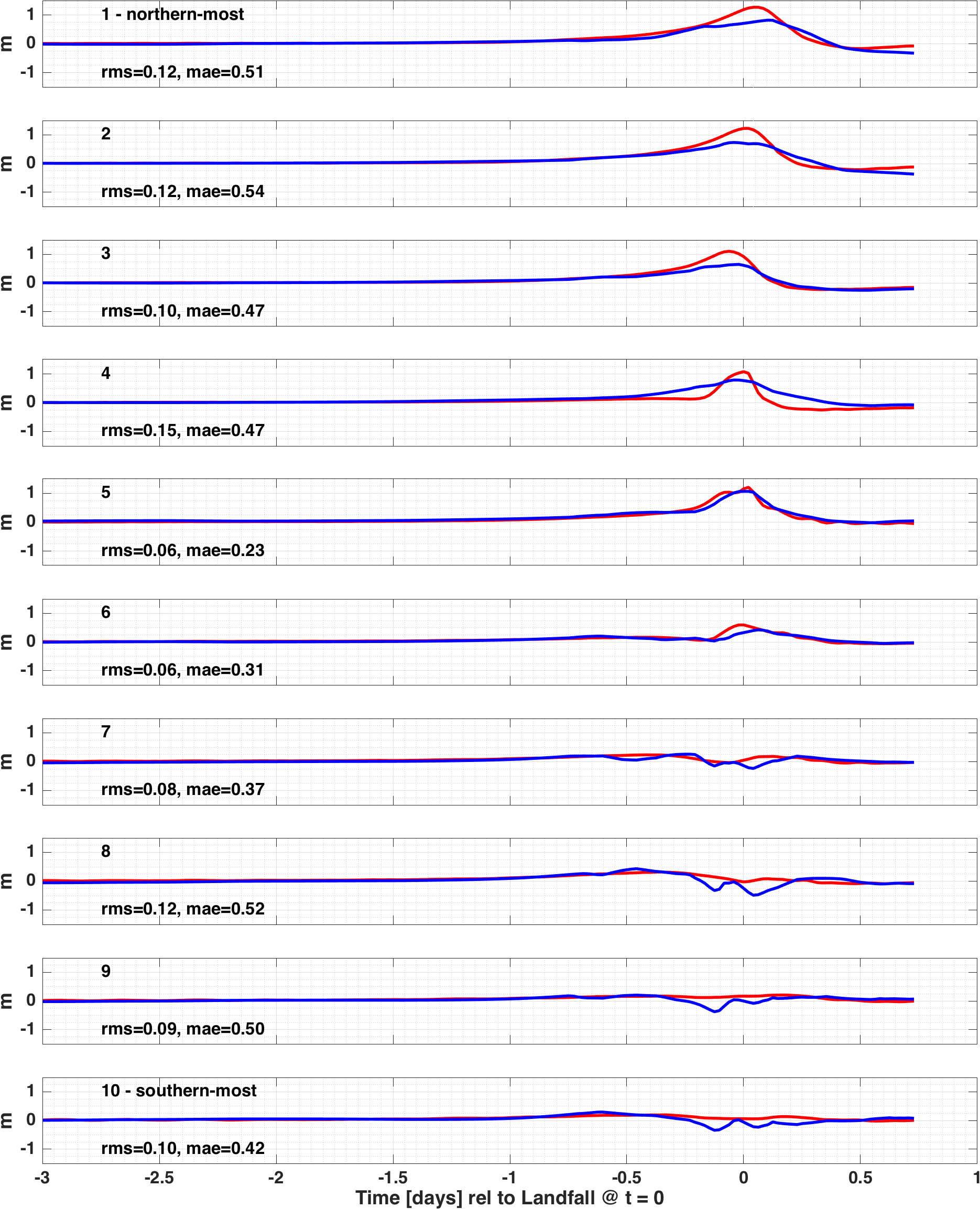}
\caption{Performance of ANN (32, 64) on Hurricane Isabel (2003). ADCIRC model results ("observations") are shown with red lines; ANN model outputs are shown with blue lines.}
\label{Fig:Isabel_performance} 
\end{figure}

Hurricane Isabel made landfall on the Outer Banks of North Carolina near Drum Inlet (between output locations 5 and 6), as a strong category 2 storm. Figure~\ref{Fig:Isabel_performance} summarizes performance of ANN (32,64) on Hurricane Isabel (2003).   Even though Isabel made landfall within the landfalling zone covered by the storms in the data set, there is a significant storm response along the northern NC coast (locations 1-4).  This is expected since the size of the storm puts hurricane-force winds onshore and well to the north of landfall.  There is very little "truth" storm surge response to the left (southwest) of the storm, but the ANN model shows negative water levels at locations 8-10.    The figure also demonstrates that ANN generally under-predicts storm surge response in the north, where the ADCIRC storm surge is largest. 

Previously, the experiments on synthetic storms also showed that locations 8-10 were the most challenging for the ANN with the greatest range of possible values. We interpret this to mean that the ANN training phase  attempts to minimize the error by increasing the weight values, which inadvertently causes non-physical (negative) values of the storm surges. Unfortunately, none of the \emph{feedforward} ANNs that were studied in this research could demonstrate exceptional accuracy at locations 8-10, which implies that the information contained in the input vector ($\vec{x}$) may be \emph{insufficient}. Had the network have some form of memory, it could have derived the missing information and use it, but this is not possible due to the limitations of the feedforward network architecture.

\section{Conclusions}
\label{sctConclusions}
The paper presents a multiple output feedforward ANN for storm surge predictions in North Carolina. Several architectures and network sizes were compared with a three layer feedforward ANN with 32 and 64 neurons being the optimal in terms of its accuracy vs. complexity. The model and its implementation is freely available and can be applied to other coastal locations, provided that sufficient training data is available.

The model accuracy at each output is comparable or slightly better than that reported by other researchers. Even though the average errors were low, the absolute predictions of the FF ANN are subject to occasional under-estimation of peak surges. Based on the number of experiments and tested architectures, we consider that this is an inherent limitation of the memory-less, feedforward ANN approach. As such, further research will target introduction of state variables or memory to the network.

\section*{Acknowledgments}
This research was performed under an appointment to the U.S. Department of Homeland Security (DHS) Science \& Technology (S\&T) Directorate Office of University Programs Summer Research Team Program for Minority Serving Institutions, administered by the Oak Ridge Institute for Science and Education (ORISE) through an interagency agreement between the U.S. Department of Energy (DOE) and DHS. ORISE is managed by ORAU under DOE contract number DE-AC05-06OR23100. All opinions expressed in this paper are the authors' and do not necessarily reflect the policies and views of DHS, DOE or ORAU/ORISE.
\bibliographystyle{elsarticle-harv}
\bibliography{storm_surge}

\begin{thebibliography}{49}
\expandafter\ifx\csname natexlab\endcsname\relax\def\natexlab#1{#1}\fi
\expandafter\ifx\csname url\endcsname\relax
  \def\url#1{\texttt{#1}}\fi
\expandafter\ifx\csname urlprefix\endcsname\relax\def\urlprefix{URL }\fi

\bibitem[{Abadi et~al.(2015)Abadi, Agarwal, Barham, Brevdo, Chen, Citro,
  Corrado, Davis, Dean, Devin, et~al.}]{abadi2015tensorflow}
Abadi, M., Agarwal, A., Barham, P., Brevdo, E., Chen, Z., Citro, C., Corrado,
  G.~S., Davis, A., Dean, J., Devin, M., et~al., 2015. Tensorflow: Large-scale
  machine learning on heterogeneous systems, 2015. Software available from
  tensorflow. org 1.

\bibitem[{Aminzadeh et~al.(2013)Aminzadeh, Sandham, and
  Leggett}]{aminzadeh2013geophysical}
Aminzadeh, F., Sandham, W., Leggett, M., 2013. Geophysical applications of
  artificial neural networks and fuzzy logic. Vol.~21. Springer Science \&
  Business Media.

\bibitem[{Anthes(1974)}]{ROG:ROG304}
Anthes, R.~A., 1974. The dynamics and energetics of mature tropical cyclones.
  Reviews of Geophysics 12~(3), 495--522.
\newline\urlprefix\url{http://dx.doi.org/10.1029/RG012i003p00495}

\bibitem[{Bajo and Umgiesser(2010)}]{bajo2010storm}
Bajo, M., Umgiesser, G., 2010. Storm surge forecast through a combination of
  dynamic and neural network models. Ocean Modelling 33~(1), 1--9.

\bibitem[{Bernier and Thompson(2015)}]{bernier2015deterministic}
Bernier, N.~B., Thompson, K.~R., 2015. Deterministic and ensemble storm surge
  prediction for atlantic canada with lead times of hours to ten days. Ocean
  Modelling 86, 114--127.

\bibitem[{Bezuglov(2016)}]{git_url}
Bezuglov, A., 2016. {ANN} for storm surge prediction.
  \url{https://github.com/abezuglov/ANN}, [Online; accessed 19-July-2016].

\bibitem[{Bishop(2006)}]{bishop2006pattern}
Bishop, C.~M., 2006. Pattern recognition. Machine Learning 128.

\bibitem[{Blanton et~al.(2012{\natexlab{a}})Blanton, Luettich, Vickery, Hanson,
  Slover, and Langan}]{Blanton2012ncfmpstats}
Blanton, B., Luettich, R., Vickery, P., Hanson, J., Slover, K., Langan, T.,
  2012{\natexlab{a}}. {North Carolina Floodplain Mapping Program: Coastal Flood
  Insurance Study - Production Simulations and Statistical Analyses}. Technical
  Report TR-12-03, Renaissance Computing Institute, The University of North
  Carolina at Chapel Hill.

\bibitem[{Blanton et~al.(2012{\natexlab{b}})Blanton, McGee, Fleming, Kaiser,
  Kaiser, Lander, Luettich, Dresback, and Kolar}]{Blanton_etal:2012}
Blanton, B., McGee, J., Fleming, J., Kaiser, C., Kaiser, H., Lander, H.,
  Luettich, R., Dresback, K., Kolar, R., 2012{\natexlab{b}}. Urgent computing
  of storm surge for {N}orth {C}arolina's coast. Procedia Computer Science
  9~(0), 1677 -- 1686, proceedings of the International Conference on
  Computational Science, ICCS 2012.
\newline\urlprefix\url{http://www.sciencedirect.com/science/article/pii/S1877050912003067}

\bibitem[{Das et~al.(2011)Das, Jung, Ebersole, Wamsley, and
  Whalin}]{das2011efficient}
Das, H.~S., Jung, H., Ebersole, B., Wamsley, T., Whalin, R.~W., 2011. An
  efficient storm surge forecasting tool for coastal mississippi. Coastal
  Engineering Proceedings 1~(32), 21.

\bibitem[{Davis et~al.(2010)Davis, Paramygin, Forrest, and
  Sheng}]{Davis_etal:10}
Davis, J.~R., Paramygin, V.~A., Forrest, D., Sheng, Y.~P., 2011/10/27 2010.
  Toward the probabilistic simulation of storm surge and inundation in a
  limited-resource environment. Mon. Wea. Rev. 138~(7), 2953--2974.
\newline\urlprefix\url{http://dx.doi.org/10.1175/2010MWR3136.1}

\bibitem[{De~Oliveira et~al.(2009)De~Oliveira, Ebecken, De~Oliveira, and
  de~Azevedo~Santos}]{de2009neural}
De~Oliveira, M.~M., Ebecken, N. F.~F., De~Oliveira, J. L.~F.,
  de~Azevedo~Santos, I., 2009. Neural network model to predict a storm surge.
  Journal of {A}pplied {M}eteorology and {C}limatology 48~(1), 143--155.

\bibitem[{Di~Liberto et~al.(2011)Di~Liberto, Colle, Georgas, Blumberg, and
  Taylor}]{di2011verification}
Di~Liberto, T., Colle, B.~A., Georgas, N., Blumberg, A.~F., Taylor, A.~A.,
  2011. Verification of a multimodel storm surge ensemble around {N}ew {Y}ork
  {C}ity and {L}ong {I}sland for the cool season. Weather and Forecasting
  26~(6), 922--939.

\bibitem[{Dietrich et~al.(2011)Dietrich, Westerink, Kennedy, Smith, Jensen,
  Zijlema, Holthuijsen, Dawson, Luettich, Powell, Cardone, Cox, Stone,
  Pourtaheri, Hope, Tanaka, Westerink, Westerink, and
  Cobell}]{Dietrich_etal:2011}
Dietrich, J.~C., Westerink, J.~J., Kennedy, A.~B., Smith, J.~M., Jensen, R.~E.,
  Zijlema, M., Holthuijsen, L.~H., Dawson, C., Luettich, R.~A., Powell, M.~D.,
  Cardone, V.~J., Cox, A.~T., Stone, G.~W., Pourtaheri, H., Hope, M.~E.,
  Tanaka, S., Westerink, L.~G., Westerink, H.~J., Cobell, Z., 2011. Hurricane
  {G}ustav (2008) waves and storm surge: Hindcast, synoptic analysis, and
  validation in southern louisiana. Monthly Weather Review 139~(8), 2488--2522.
\newline\urlprefix\url{http://dx.doi.org/10.1175/2011MWR3611.1}

\bibitem[{Dresback et~al.(2013)Dresback, Fleming, Blanton, Kaiser, Gourley,
  Tromble, Jr., Kolar, Hong, Cooten, Vergara, Flamig, Lander, Kelleher, and
  Nemunaitis-Monroe}]{Dresback_etal:2013}
Dresback, K., Fleming, J., Blanton, B., Kaiser, C., Gourley, J., Tromble, E.,
  Jr., R.~L., Kolar, R., Hong, Y., Cooten, S., Vergara, H., Flamig, Z., Lander,
  H., Kelleher, K., Nemunaitis-Monroe, K., 2013. Skill assessment of a
  real-time forecast system utilizing a coupled hydrologic and coastal
  hydrodynamic model during {H}urricane {I}rene (2011). Cont. Shelf Res. 71,
  78--94.

\bibitem[{Flather(2001)}]{Flather:2001}
Flather, R., 2001. Storm surges. In: Steele, J., Thorpe, S., Turekian, K.
  (Eds.), Encyclopedia of Ocean Science. Academic Press, pp. 2882--2892.

\bibitem[{Govindaraju and Rao(2013)}]{govindaraju2013artificial}
Govindaraju, R.~S., Rao, A.~R., 2013. Artificial neural networks in hydrology.
  Vol.~36. Springer Science \& Business Media.

\bibitem[{Hashemi et~al.(2016)Hashemi, Spaulding, Shaw, Farhadi, and
  Lewis}]{hashemi2016efficient}
Hashemi, M.~R., Spaulding, M.~L., Shaw, A., Farhadi, H., Lewis, M., 2016. An
  efficient artificial intelligence model for prediction of tropical storm
  surge. Natural Hazards 82~(1), 471--491.

\bibitem[{Holland(1980)}]{holland1980analytic}
Holland, G.~J., 1980. An analytic model of the wind and pressure profiles in
  hurricanes. Monthly Weather Review 108~(8), 1212--1218.

\bibitem[{Holland(2008)}]{holland2008revised}
Holland, G.~J., 2008. A revised hurricane pressure-wind model. Monthly Weather
  Review 136~(9), 3432--3445.

\bibitem[{Holland et~al.(2010)Holland, Belanger, and
  Fritz}]{holland2010revised}
Holland, G.~J., Belanger, J.~I., Fritz, A., 2010. A revised model for radial
  profiles of hurricane winds. Monthly Weather Review 138~(12), 4393--4401.

\bibitem[{Jelesnianski et~al.(1992)Jelesnianski, Chen, and
  Shaffer}]{Jelesnianski_etal:1992}
Jelesnianski, C., Chen, J., Shaffer, W.~A., 1992. {SLOSH: Sea, Lake, and
  Overland Surges from Hurricanes.} NOAA Technical Report NWS 48, U.S.
  Department of Commerce.

\bibitem[{Jia and Taflanidis(2013)}]{jia2013kriging}
Jia, G., Taflanidis, A.~A., 2013. Kriging metamodeling for approximation of
  high-dimensional wave and surge responses in real-time storm/hurricane risk
  assessment. Computer Methods in Applied Mechanics and Engineering 261,
  24--38.

\bibitem[{Khemchandani et~al.(2009)Khemchandani, Chandra,
  et~al.}]{khemchandani2009regularized}
Khemchandani, R., Chandra, S., et~al., 2009. Regularized least squares fuzzy
  support vector regression for financial time series forecasting. Expert
  Systems with Applications 36~(1), 132--138.

\bibitem[{Khuri and Mukhopadhyay(2010)}]{WICS:WICS73}
Khuri, A.~I., Mukhopadhyay, S., 2010. Response surface methodology. Wiley
  Interdisciplinary Reviews: Computational Statistics 2~(2), 128--149.
\newline\urlprefix\url{http://dx.doi.org/10.1002/wics.73}

\bibitem[{Kim et~al.(2015)Kim, Melby, Nadal-Caraballo, and
  Ratcliff}]{kim2015time}
Kim, S.-W., Melby, J.~A., Nadal-Caraballo, N.~C., Ratcliff, J., 2015. A
  time-dependent surrogate model for storm surge prediction based on an
  artificial neural network using high-fidelity synthetic hurricane modeling.
  Natural Hazards 76~(1), 565--585.

\bibitem[{Kingma and Ba(2014)}]{kingma2014adam}
Kingma, D., Ba, J., 2014. Adam: A method for stochastic optimization. arXiv
  preprint arXiv:1412.6980.

\bibitem[{Knapp et~al.(2010)Knapp, Kruk, Levinson, Diamond, and
  Neumann}]{Knapp_etal:2010}
Knapp, K., Kruk, M., Levinson, D., Diamond, H., Neumann, C., 2010. {The
  International Best Track Archive for Climate Stewardship (IBTrACS): U}nifying
  tropical cyclone best track data. Bulletin of the American Meteorological
  Society 91, 363--376.

\bibitem[{Lee(2006)}]{lee2006neural}
Lee, T.-L., 2006. Neural network prediction of a storm surge. Ocean Engineering
  33~(3), 483--494.

\bibitem[{Lee(2008)}]{lee2008back}
Lee, T.-L., 2008. Back-propagation neural network for the prediction of the
  short-term storm surge in {T}aichung {H}arbor, {T}aiwan. Engineering
  Applications of Artificial Intelligence 21~(1), 63--72.

\bibitem[{Lin et~al.(2010)Lin, Smith, Villarini, Marchok, and
  Baeck}]{Lin_etal:2010a}
Lin, N., Smith, J.~A., Villarini, G., Marchok, T.~P., Baeck, M.~L., 2011/10/27
  2010. Modeling extreme rainfall, winds, and surge from hurricane isabel
  (2003). Weather and Forecasting 25~(5), 1342--1361.
\newline\urlprefix\url{http://dx.doi.org/10.1175/2010WAF2222349.1}

\bibitem[{Luettich~Jr et~al.(1992)Luettich~Jr, Westerink, and
  Scheffner}]{luettich1992adcirc}
Luettich~Jr, R., Westerink, J., Scheffner, N.~W., 1992. {ADCIRC}: An advanced
  three-dimensional circulation model for shelves, coasts, and estuaries.
  report 1. theory and methodology of adcirc-2ddi and adcirc-3dl. Tech. rep.,
  DTIC Document.

\bibitem[{Ma et~al.(2003)Ma, Theiler, and Perkins}]{ma2003accurate}
Ma, J., Theiler, J., Perkins, S., 2003. Accurate on-line support vector
  regression. Neural computation 15~(11), 2683--2703.

\bibitem[{Marks(2003)}]{marks2003hurricanes}
Marks, F.~D., 2003. Hurricanes. Handbook of Weather, Climate, and Water:
  Dynamics, Climate, Physical Meteorology, Weather Systems, and Measurements,
  641--675.

\bibitem[{Qi et~al.(2009)Qi, Chen, Beardsley, Perrie, and Cowles}]{FVCOM_1}
Qi, J., Chen, C., Beardsley, R., Perrie, W., Cowles, G., 2009. An
  unstructured-grid finite-volume surface wave model {(FVCOM-SWAVE)}:
  {I}mplementation, validations and applications. Ocean Modelling 28, 153--166.

\bibitem[{Rajasekaran et~al.(2008)Rajasekaran, Gayathri, and
  Lee}]{rajasekaran2008support}
Rajasekaran, S., Gayathri, S., Lee, T.-L., 2008. Support vector regression
  methodology for storm surge predictions. Ocean Engineering 35~(16),
  1578--1587.

\bibitem[{Resio and Westerink(2008)}]{ResioWesterink:08}
Resio, D., Westerink, J., September 2008. Modeling the physics of storm surges.
  Physics Today.

\bibitem[{Rumelhart et~al.(1988)Rumelhart, Hinton, and
  Williams}]{rumelhart1988learning}
Rumelhart, D.~E., Hinton, G.~E., Williams, R.~J., 1988. Learning
  representations by back-propagating errors. Cognitive modeling 5~(3), 1.

\bibitem[{Scott(2015)}]{scott2015multivariate}
Scott, D.~W., 2015. Multivariate density estimation: theory, practice, and
  visualization. John Wiley \& Sons.

\bibitem[{Skiena(1998)}]{skiena1998algorithm}
Skiena, S.~S., 1998. The algorithm design manual: Text. Vol.~1. Springer
  Science \& Business Media.

\bibitem[{Smola and Vapnik(1997)}]{smola1997support}
Smola, A., Vapnik, V., 1997. Support vector regression machines. Advances in
  neural information processing systems 9, 155--161.

\bibitem[{Sreekumar et~al.(2015)Sreekumar, Verma, Sujil, and
  Kumar}]{sreekumar2015one}
Sreekumar, S., Verma, J., Sujil, A., Kumar, R., 2015. One day forth forecasting
  of hourly electrical load using genetically tuned support vector regression
  for smart grid frame work. In: 2015 2nd International Conference on Recent
  Advances in Engineering \& Computational Sciences (RAECS). IEEE, pp. 1--6.

\bibitem[{Taflanidis et~al.(2013)Taflanidis, Kennedy, Westerink, Smith, Cheung,
  Hope, and Tanaka}]{taflanidis2013rapid}
Taflanidis, A.~A., Kennedy, A.~B., Westerink, J.~J., Smith, J., Cheung, K.~F.,
  Hope, M., Tanaka, S., 2013. Rapid assessment of wave and surge risk during
  landfalling hurricanes: probabilistic approach. Journal of Waterway, Port,
  Coastal, and Ocean Engineering 139~(3), 171--182.

\bibitem[{Vapnik et~al.(1997)Vapnik, Golowich, Smola,
  et~al.}]{vapnik1997support}
Vapnik, V., Golowich, S.~E., Smola, A., et~al., 1997. Support vector method for
  function approximation, regression estimation, and signal processing.
  Advances in Neural Information Processing Systems, 281--287.

\bibitem[{Westerink et~al.(2008)Westerink, Luettich, Feyen, Atkinson, Dawson,
  Roberts, Powell, Dunion, Kubatko, and Pourtaheri}]{Westerink_etal:08}
Westerink, J., Luettich, R., Feyen, J., Atkinson, J., Dawson, C., Roberts, H.,
  Powell, M., Dunion, J., Kubatko, E., Pourtaheri, H., 2008. A basin- to
  channel-scale unstructured grid hurricane storm surge model applied to
  {S}outhern {L}ouisiana. Mon. Weather Rev. 136, 833--864.

\bibitem[{Wolff et~al.(2016)Wolff, K{\"u}hnert, Lorenz, Kramer, and
  Heinemann}]{wolff2016comparing}
Wolff, B., K{\"u}hnert, J., Lorenz, E., Kramer, O., Heinemann, D., 2016.
  Comparing support vector regression for pv power forecasting to a physical
  modeling approach using measurement, numerical weather prediction, and cloud
  motion data. Solar Energy 135, 197--208.

\bibitem[{Yu et~al.(2006)Yu, Chen, and Chang}]{yu2006support}
Yu, P.-S., Chen, S.-T., Chang, I.-F., 2006. Support vector regression for
  real-time flood stage forecasting. Journal of Hydrology 328~(3), 704--716.

\bibitem[{Zhang et~al.(1998)Zhang, Patuwo, and Hu}]{zhang1998forecasting}
Zhang, G., Patuwo, B.~E., Hu, M.~Y., 1998. Forecasting with artificial neural
  networks:: The state of the art. International journal of forecasting 14~(1),
  35--62.

\bibitem[{Zhang et~al.(2006)Zhang, Sato, and Iai}]{zhang2006support}
Zhang, J., Sato, T., Iai, S., 2006. Support vector regression for on-line
  health monitoring of large-scale structures. Structural Safety 28~(4),
  392--406.

\end{thebibliography}

\end{document}